\documentclass[conference]{IEEEtran}
\IEEEoverridecommandlockouts
\usepackage{cite}
\usepackage{amsmath,amssymb,amsfonts}
\usepackage{graphicx}
\graphicspath{{./images/}}
\usepackage{textcomp}
\usepackage{xcolor}
\usepackage{makecell}
\usepackage{gensymb}
\usepackage{multirow}
\usepackage{amssymb}% http://ctan.org/pkg/amssymb
\usepackage{pifont}% http://ctan.org/pkg/pifont
\usepackage{caption}
\usepackage{subcaption}
\usepackage{algorithm}
\usepackage{algpseudocode}

\def\BibTeX{{\rm B\kern-.05em{\sc i\kern-.025em b}\kern-.08em
		T\kern-.1667em\lower.7ex\hbox{E}\kern-.125emX}}

\usepackage{tikz,xcolor,hyperref}
\definecolor{lime}{HTML}{A6CE39}
\DeclareRobustCommand{\orcidicon}{
	\begin{tikzpicture}
		\draw[lime, fill=lime] (0,0) 
		circle [radius=0.16] 
		node[white] {{\fontfamily{qag}\selectfont \tiny ID}};
		\draw[white, fill=white] (-0.0625,0.095) 
		circle [radius=0.007];
	\end{tikzpicture}
	\hspace{-2mm}
}

\foreach \x in {A,B}{\expandafter\xdef\csname orcid\x\endcsname{\noexpand\href{https://orcid.org/\csname orcidauthor\x\endcsname}
		{\noexpand\orcidicon}}
}

\newcommand\copyrighttext{%
	\footnotesize © 2022 IEEE. Personal use of this material is permitted. Permission from IEEE must be 
	obtained for all other uses, in any current or future media, including 
	reprinting/republishing this material for advertising or promotional purposes, creating new 
	collective works, for resale or redistribution to servers or lists, or reuse of any copyrighted 
	component of this work in other works. \href{dx.doi.org/10.1109/IV51971.2022.9827458}{DOI:10.1109/IV51971.2022.9827458} }
\newcommand\copyrightnotice{%
	\begin{tikzpicture}[remember picture,overlay]
		\node[anchor=south,yshift=10pt] at (current page.south) {\fbox{\parbox{\dimexpr\textwidth-\fboxsep-\fboxrule\relax}{\copyrighttext}}};
	\end{tikzpicture}%
}

\begin{document}
	
	\title{Uncertainty-Aware Prediction of Battery Energy Consumption for Hybrid Electric Vehicles}
	
	\author{Jihed Khiari  \orcidA{} \IEEEmembership {Student Member, IEEE} and Cristina~Olaverri-Monreal \orcidB{}~\IEEEmembership{Senior Member, IEEE}% <-this stops a space
		
		\thanks{ITS-Chair for Sustainable Transport Logistics 4.0., Johannes Kepler University Linz, Austria. \{jihed.khiari, cristina.olaverri-monreal\}@jku.at  }

		% <-this stops a space

		%\title{Estimating Trip Efficiency: \\ an Unsupervised %Learning Approach \\
		}
		
		%	\author{\IEEEauthorblockN{Jihed Khiari \orcidA{}}
			%		\IEEEauthorblockA{\textit{Chair for ITS-Sustainable %Transport Logistics 4.0} \\
				%			\textit{ Johannes Kepler University }\\
				%			Linz, Austria \\
				%			jihed.khiari@jku.at}
			%
			%		\and
			%		\IEEEauthorblockN{Cristina Olaverri-Monreal \orcidB{}}
			%		\IEEEauthorblockA{\textit{Chair for ITS-Sustainable %Transport Logistics 4.0} \\
				%		\textit{ Johannes Kepler University}\\
				%		Linz, Austria \\
				%		cristina.olaverri-monreal@jku.at }
			%}

		\maketitle
		\copyrightnotice
		
		\begin{abstract}
			The usability of vehicles is highly dependent on their energy consumption. In particular, one of the main factors hindering the mass adoption of electric (EV), hybrid (HEV), and plug-in hybrid (PHEV) vehicles is \textit{range anxiety}, which occurs when a driver is uncertain about the availability of energy for a given trip. To tackle this problem, we propose a machine learning approach for modeling the battery energy consumption. By reducing predictive uncertainty, this method can help increase trust in the vehicle's performance and thus boost its usability. Most related work focuses on physical and/or chemical models of the battery that affect the energy consumption. We propose a data-driven approach which relies on real-world datasets including battery related attributes. Our approach showed an improvement in terms of predictive uncertainty as well as in accuracy compared to traditional methods.
			
		\end{abstract}
		
		\begin{IEEEkeywords}
			hybrid vehicle, energy range, uncertainty-aware prediction, battery energy
		\end{IEEEkeywords}
		
		\section{Introduction}
		Battery-equipped vehicles such as electric, hybrid, and plug-in hybrid vehicles can contribute to a reduction in transport-related greenhouse gas emissions, since they can rely on electric energy rather than or in addition to fuel~\cite{malikopoulos2013impact}~\cite{pistoia2010electric}. However, the limited charging infrastructure and battery capacity restrict their usability~\cite{gnann2018fast}~\cite{conte2006battery}. Furthermore, the estimation of the current vehicle range lacks accuracy and trustworthiness~\cite{franke2015advancing}~\cite{birrell2014defining}. The combination of these factors can lead to \textit{range anxiety}; the idea that the currently available energy might be insufficient to reach the planned destination, thus causing the driver to feel anxious and uncertain. To alleviate this problem, a measure of uncertainty about the range prediction can be displayed to the driver. Studies have shown that this would improve the driver experience and trust in the displayed values~\cite{jung2015displayed}~\cite{eisel2016understanding}. 
		
		In the literature,  many approaches have been proposed to tackle range estimation, e.g. using model-based approaches~\cite{fiori2020energy}~\cite{hu2020reliable}~\cite{oliva2013model}. On the other hand, it is crucial to note that several factors can affect the battery energy consumption of a vehicle: the chosen route, the driving area topography, the driver behavior (e.g. accelerating/decelerating patterns), the effects of regenerative braking, the battery type and age, sensor noise, whether or not fuel is used alongside the battery, etc. Therefore, the battery's energy consumption is subject to uncertainty and variability. While there are available methods tackling such uncertainty and its various sources such as ~\cite{jung2015displayed}~\cite{fiori2020energy}~\cite{sautermeister2017influence}~\cite{pelletier2019electric}, most of these works consider purely electric vehicles.
		
\begin{figure}
			\begin{subfigure}{0.47\textwidth}
				
				\includegraphics[width=\textwidth]{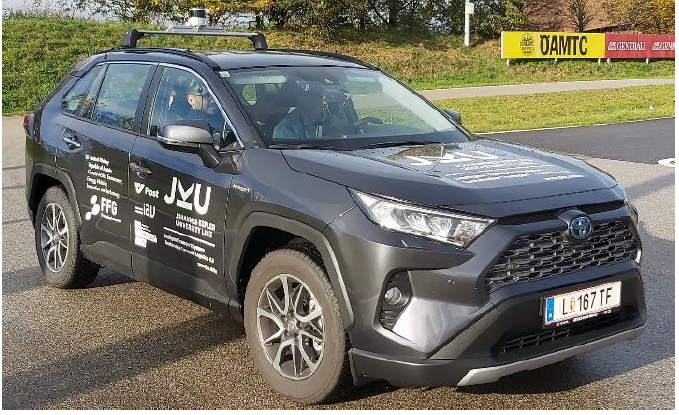}
				\caption{JKU-ITS vehicle used for data collection}
				\label{fig:car}
			\end{subfigure}
			\begin{subfigure}{0.47\textwidth}
				\includegraphics[width=\textwidth]{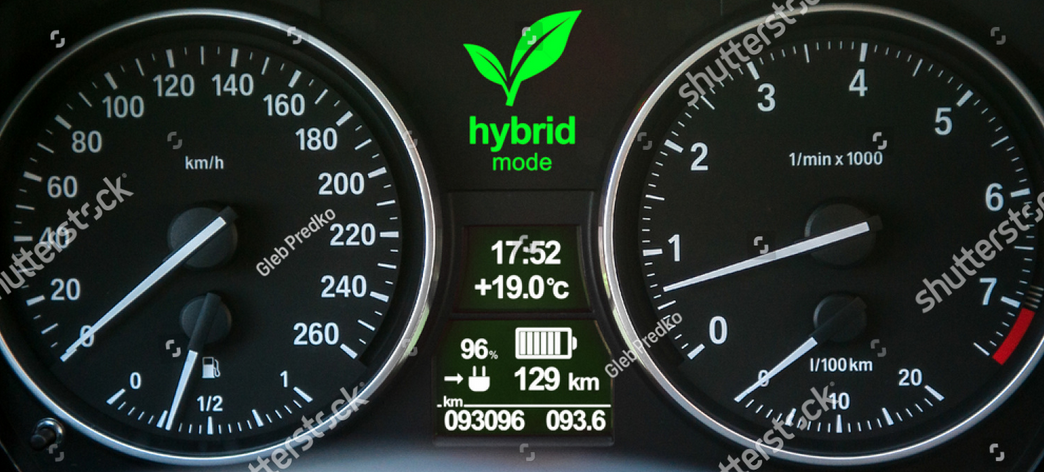}
				\caption{Dashboard of a hybrid electric vehicle}
				\label{fig:dash}
			\end{subfigure}
			
			\caption{The JKU-ITS Vehicle and a typical dashboard}	
		\end{figure}

		Since fuel availability and utilization can affect the energy consumption pattern, we focus in this study on predicting battery usage for hybrid electric vehicles, which currently have a globally higher market share than electric vehicles~\cite{palmer2018total}. Nonetheless, the approach remains applicable to any battery-equipped vehicle. To further extend the state of the art, we propose to utilize a method that estimates predictive uncertainty using a deep neural networks (DNN) ensemble~\cite{lakshminarayanan2017simple}. Our approach is based on data that we collected using the JKU-ITS vehicle shown in~\ref{fig:car} and described in~\cite{certad2022}. Fig.~\ref{fig:dash} depicts a typical hybrid vehicle dashboard where the battery state of charge is shown in terms of a percentage, and is also illustrated in a battery pictogram. We propose to provide the driver with more information regarding the battery state of charge, namely a prediction that is generated in an uncertainty-aware fashion. Moreover, it would be possible to output not only the predicted value, but also the associated variance, thus depicting the level of confidence surrounding the prediction. To do so, we use a DNN ensemble that minimizes predictive uncertainty. Therefore, it can be beneficial in accurately and reliably informing the driver about the battery state. For instance, this approach can be integrated in a human-machine interface or used offline, for instance; to analyze the battery energy consumption in historical trips, or forecast it for future trips. \\
		
		The remainder of this paper is structured as follows. The next section considers related work in the field of uncertainty estimation for the battery usage of electric vehicles. Section~\ref{sec:method} describes the steps that were
		followed to implement our approach. Section~\ref{sec:results} presents the results regarding the predicted energy values. Finally Sections~\ref{sec:discussion} and~\ref{sec:conclusion} discuss the findings and conclude the paper.

		\section{Related Work}
		\label{sec:rel_work}
		Due to the high variability of battery energy consumption, there are numerous available methods for battery range prediction that focus on electric vehicles. Most of these methods rely on kinetic, physical, and/or chemical features of the vehicle battery to estimate the range and uncertainty such as~\cite{oliva2013model} and~\cite{sautermeister2017influence}. In~\cite{oliva2013model}, the authors propose a model-based approach for predicting the remaining driving range of an electric vehicle by combining a particle filter with Markov chains. In~\cite{sautermeister2017influence}, the authors examine the uncertainty of range estimation, by estimating the probability of reaching a given destination on a given route. Similarly, ~\cite{ondruska2014probabilistic} proposes a probabilistic method that outputs confidence levels in reaching a destination based on its location and the possible trajectory segments.
		
		On the other hand, several works focus on routing solutions under uncertainty and/or maximizing range utilization such as~\cite{pelletier2019electric},~\cite{thibault2018unified},~\cite{bedogni2014driving}, for instance, by taking into account eco-driving techniques or robust optimization solvers. 
		
		Works such as~\cite{jung2015displayed} emphasize the importance of visualizing energy availability (e.g. SOC) values in terms of ranges rather than deterministic single values that are highly likely to be wrong, hence eroding the trust of the user in these values. To do so, the authors conduct a field study where they investigate the effect of displayed battery information on electric vehicle drivers. Moreover, several familiar examples are stated such as a smartphone displaying a battery availability of 20\% only to shut down quickly thereafter. In fact, the user experience improves considerably when the displayed values tend to match the reality. With the increasing reliance on battery powered devices in crucial areas like hospitals and public roads (e.g. battery equipped vehicles), it is all the more important to have consistent and robust outputs from these devices, e.g. their state of charge.
		
		It is also noteworthy that there can be a considerable difference of the projected energy consumption by a traffic simulator and the actual one~\cite{validi2021analysis}. Similarly, the vehicle manufacturers typically provide energy consumption values that are overly optimistic~\cite{de2015energy}. The utilization conditions vary considerably, and this can be reflected in driving data. Therefore, we seek to rely solely on recorded battery energy consumption data acquired from driving vehicles. Furthermore, we will focus in our study on hybrid electric vehicles. To this end, we are applying deep neural network ensemble method (or deep ensemble) that specifically targets predictive uncertainty.
		
		Deep neural networks have been established as powerful prediction methods with wide successes in fields such as computer vision, natural language processing, and robotics~\cite{lecun2015deep}~\cite{zhang2018survey}~\cite{sunderhauf2018limits}. Nevertheless, they are prone to possibly producing wrong results with high confidence, i.e. not knowing when they do not know, thus posing security risks~\cite{amodei2016concrete}~\cite{carlini2017towards}. This has led to an increase in research about adversarial training \cite{kurakin2016adversarial} that is designed to increase robustness. As for quantifying uncertainty, the most common approach, so far, is to use Bayesian networks~\cite{bernardo2009bayesian}, which are complex to implement and computationally expensive. That's why, it is highly relevant to utilize a simpler method such as the DNN ensemble described in~\cite{lakshminarayanan2017simple}. The use of ensembles to improve predictive performance is relatively common~\cite{dietterich2000ensemble}~\cite{mendes2012ensemble}. In~\cite{lakshminarayanan2017simple}, it is shown that they can also be utilized to quantify predictive uncertainty. To the authors' knowledge, a similar method has yet to be used in the context of vehicle energy consumption prediction. We therefore investigate in this work its usability and potential for providing robust and uncertainty-aware predictions. \\

		\section{Methodology}
		
		\label{sec:method}
		\subsection{DNN Ensemble}
		
		%% integrate this section in the following
		%Thereby, we can output battery state of charge (SOC) values to the driver with a reduced predictive uncertainty. With this goal in mind, we rely on data that we collected using the JKU-ITS vehicle; a hybrid electric vehicle undertaking trips around the area of Linz, Austria~\cite{certad2022}.  Given the available data, we focus on common attributes related to a vehicle's battery and show how the DNN ensemble can be beneficial in accurately and reliably informing the driver about the battery state. 	
		%%
		Our goal consisted in outputting the battery state of charge (SOC) values to the driver with a reduced predictive uncertainty. To do so, we relied on data that we collected using the JKU-ITS vehicle; a hybrid electric vehicle undertaking trips around the area of Linz, Austria~\cite{certad2022}.  Given the available data, we focused on common attributes related to a vehicle's battery and showed how the DNN ensemble can be used to predict the battery's state of charge in an uncertainty-aware fashion. The problem can be fomulated as follows. 	\\

		As detailed in~\cite{lakshminarayanan2017simple}, we considered a dataset $D$, such as $D = \{x_n,y_n\}_{n=1}^N$, where $x \in \mathbb{R}^d$ stands for the d-dimensional features, while  $y \in \mathbb{R}$ is the prediction target. The method is based on a neural network which models the probabilistic predictive distribution $p_\theta(y|x)$ where $\theta$ are the parameters of the neural network. The method relies on two key aspects:
		
		\begin{enumerate}
			\item Using a proper scoring rule as the training criterion
			%\item Using adversarial training to smooth the predictive distributions
			\item Training an ensemble of $M$ neural networks \\
		\end{enumerate}
		
		First, a scoring rule  reflects, by definition, the quality of predictive uncertainty and rewards better calibrated predictions. A review of scoring rules can be found in~\cite{gneiting2007strictly}. Typically, neural networks used for regression output a single value and the parameters are optimized to minimize the mean squared error (MSE) given by Equation~\ref{eq:mse}. As in~\cite{nix1994estimating}, we used a network that outputs two values in the final layer: the predicted mean and variance by treating the observed value as a sample from a Gaussian distribution. With the predicted mean and variance, we minimize the negative log-likelihood criterion expressed in Equation~\ref{eq:nll}.
		
		%Second, for further simplicity, we did not perform the adversarial training in our experiments, as it is an optional step and the results were satisfactory notwithstanding.
		
		Second, we trained an ensemble of neural networks independently in parallel. When training each network, we used random batches (subsamples) of the data at each iteration, as well as a random initialization of the parameters. While the batches were drawn at random from the whole training set, the chronological order was later restored to take into account the original succession of the values. The resulting ensemble is a uniformly-weighted mixture model and combines the predictions as expressed in Equation~\ref{eq:ensemble}: 
		%$p(y|x)=M^{-1} \sum_{m=1}^{M} p_{\theta_m}(y|x,\theta_m)$. \\
		
		\begin{equation}
			\label{eq:ensemble}
			p(y|x)=M^{-1} \sum_{m=1}^{M} p_{\theta_m}(y|x,\theta_m)
		\end{equation}
		
		We used $M=5$ neural networks within the ensemble, a batch size of 100 and the Adam optimizer~\cite{zhang2018improved} with a fixed learning rate of 0.1 in our experiments. To tune the learning rate, we used a grid search on a log scale from 0.1 to $10^-5$. We performed a 80\%-20\% split between training and testing datasets and set the number of iterations to 20000.

		\begin{equation}
			\label{eq:mse}
			\sum_{n=1}^{N} (y_n-\mu(x_n))^2
		\end{equation}
		
		\begin{equation}
			\label{eq:nll}
			- \log p_{\theta}(y_n|x_n) = \frac{\log \sigma_\theta^2}{2}+ 
			\frac{(y-\mu_\theta(x))^2}{2\sigma_\theta^2(x)} + constant
		\end{equation}
		
		Contrary to~\cite{lakshminarayanan2017simple}, we did not consider the intermediary and optional step consisting in using adversarial training to smooth the predictive distributions. This was motivated by providing further simplicity to the method, thus facilitating deployment, by reducing computation time.
		\subsection{Data description}  
		
%		\begin{figure}
%			\begin{subfigure}{0.47\textwidth}
%				
%				\includegraphics[width=\textwidth]{car_jku}
%				\caption{JKU-ITS vehicle used for data collection}
%				\label{fig:car}
%			\end{subfigure}
%			\begin{subfigure}{0.47\textwidth}
%				\includegraphics[width=\textwidth]{dashboard}
%				\caption{Dashboard of a hybrid electric vehicle}
%				\label{fig:dash}
%			\end{subfigure}
%			
%			\caption{The JKU-ITS Vehicle and a typical dashboard}	
%		\end{figure}
		
		%
		%	\begin{figure}[t!]
			%		\caption{JKU-ITS vehicle used for data collection}
			%		\includegraphics[width=0.5\textwidth]{car_jku}
			%		\label{fig:car}
			%	\end{figure}
		%	
		%	\begin{figure}[h!]
			%		\caption{Dashboard of a hybrid electric vehicle}
			%		\includegraphics[width=0.5\textwidth]{dashboard}
			%		\label{fig:dash}
			%	\end{figure}
		%	

		To collect real-world driving data from a battery equipped vehicle, we used the JKU-ITS vehicle~\cite{certad2022}: a 2020 Toyota RAV4 Hybrid LE AWD shown in Fig.~\ref{fig:car}. Its battery is characterized by the aspects below: 
		
		\begin{itemize}
			\item \textbf{Maximum voltage:} 244.8 V
			\item \textbf{Capacity:} 6.5 Ah
			\item \textbf{Power:} 88 W \\
			
		\end{itemize}
		
		The datasets corresponded to six  trips in the Upper Austria region. They lasted 40-50 minutes each. The first two occurred in October 2021, while the remaining four trips took place in December 2021 in a colder environment under snowy conditions. Besides the difference in weather conditions and trajectories, we also considered varying the driver during the data collection process. Four drivers undertook the six trips. \\
		
		Given that the battery's SOC is immediately available to the driver (similarly to the fuel tank status), we consider it as our prediction target. Using the on-board diagnostics (OBD) system, we collected the following attributes: 
		
		\begin{itemize}
			\item   Vehicle speed (mph)
			\item 	Hybrid/EV Battery Power (hp)
			\item 	Hybrid/EV Battery System Current (A)
			\item 	Hybrid/EV Battery System Voltage (V)
			\item 	Hybrid Battery Temperature 1 (℃)
			\item 	Hybrid Battery Temperature 2 (℃)
			\item 	Hybrid Battery Temperature 3 (℃)
			\item 	Hybrid Battery 1 Fan (\%)
			\item 	Hybrid Battery SOC (\%) \\
		\end{itemize}

		We note that the three hybrid battery temperature attributes correspond to three distinct sensors. The data granularity is of 0.1 seconds. For each of the datasets, we applied the method described above under the same settings and hyperparameters. For computation, we used an AWS EC2 Inf1 Instance~\cite{aws_instance} which is optimized for machine learning inference applications. As a framework, we used Tensorflow~\cite{dillon2017tensorflow}.
		
		\section{Results}
		
		\label{sec:results}
		\begin{figure*}
			\centering
			\begin{subfigure}{0.33\textwidth}
				\includegraphics[width=\textwidth]{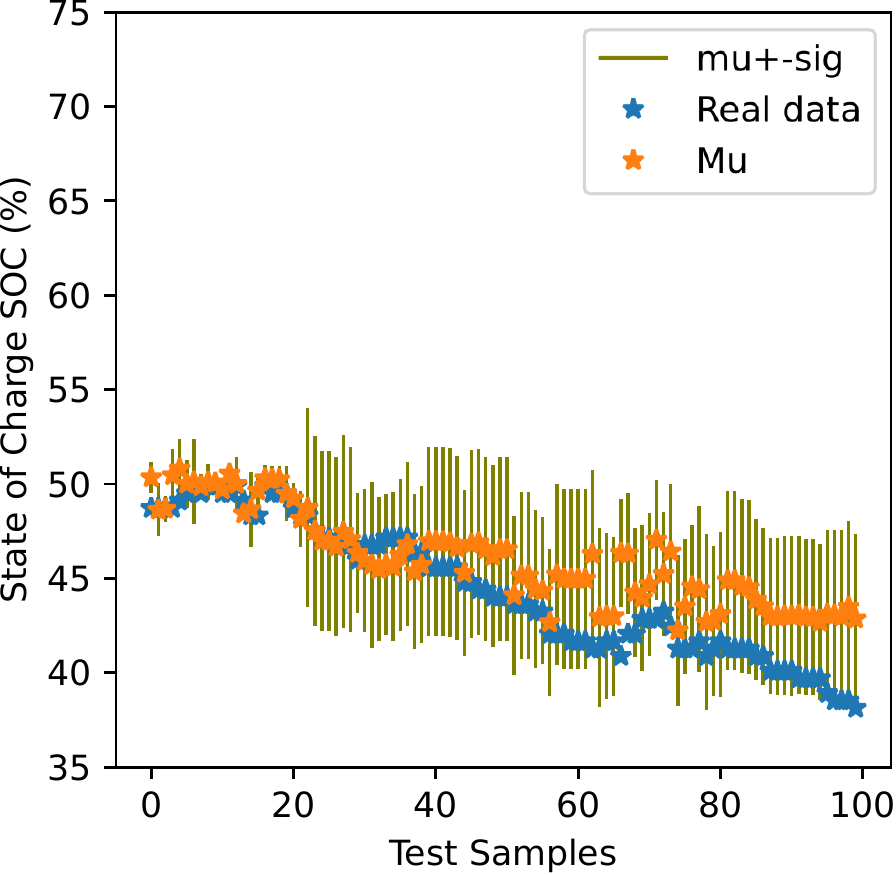}
				\caption{$D_0$}
				\label{fig:d0}
			\end{subfigure}
			\hfill
			\begin{subfigure}{0.32\textwidth}
				\includegraphics[width=\textwidth]{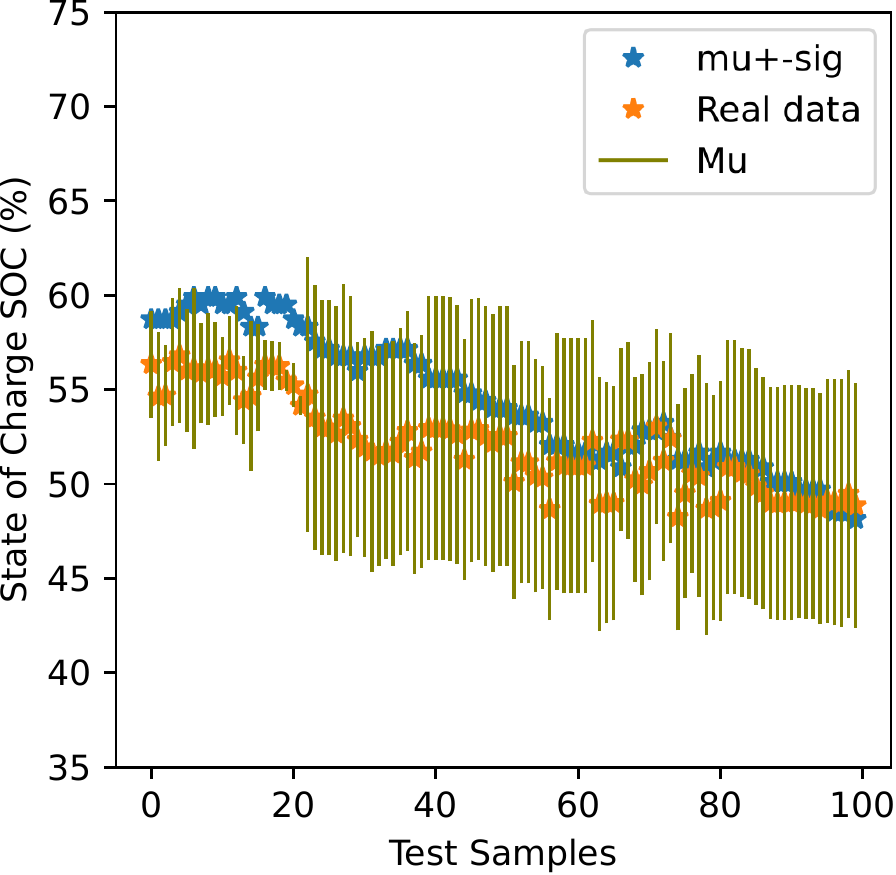}
				\caption{$D_1$}
				\label{fig:d1}
			\end{subfigure}
			\hfill
			\begin{subfigure}{0.32\textwidth}
				\includegraphics[width=\textwidth]{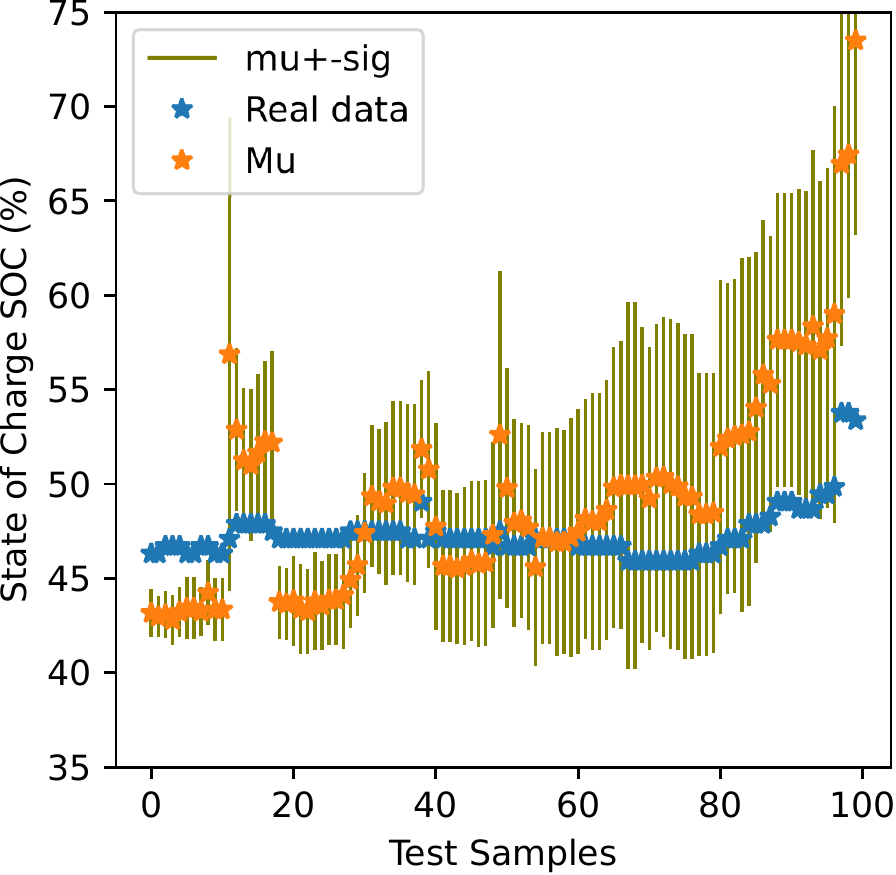}
				\caption{$D_2$}
				\label{fig:d2}
			\end{subfigure}
			
			\begin{subfigure}{0.32\textwidth}
				\includegraphics[width=\textwidth]{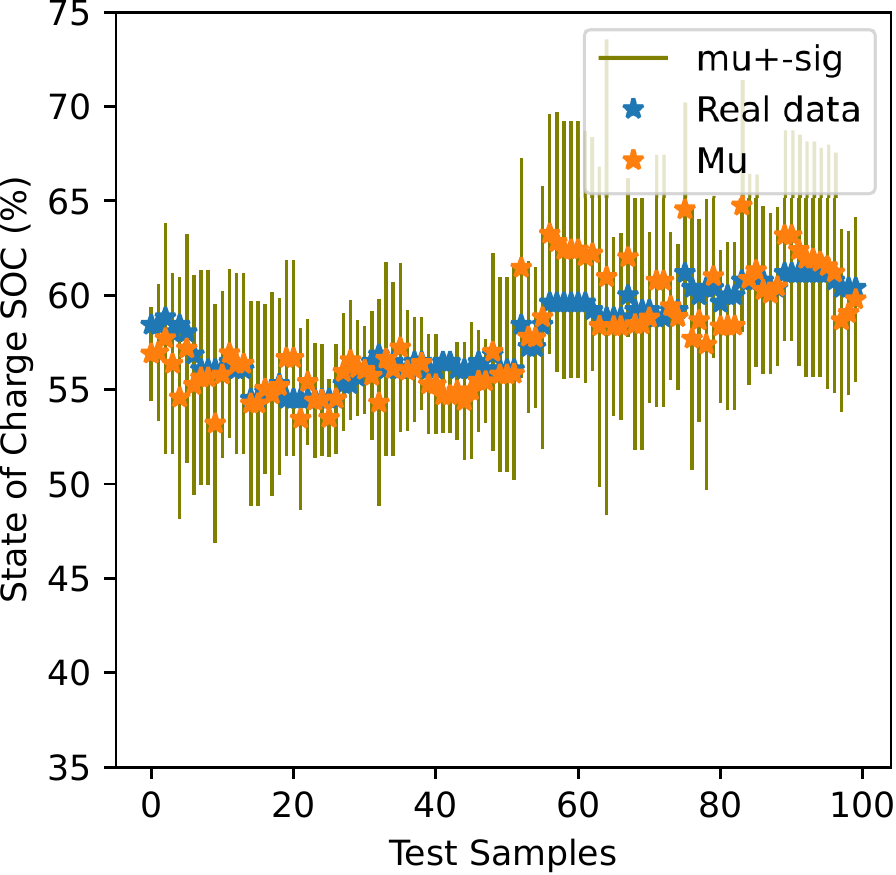}
				\caption{$D_3$}
				\label{fig:d3}
			\end{subfigure}
			\hfill
			\begin{subfigure}{0.32\textwidth}
				\includegraphics[width=\textwidth]{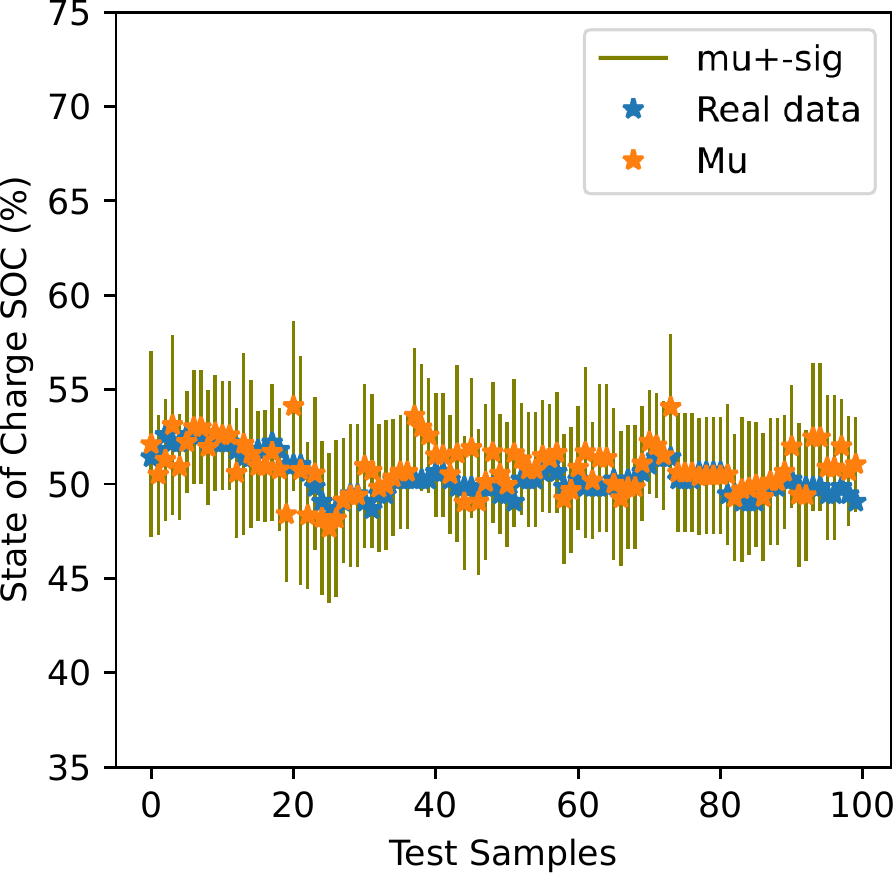}
				\caption{$D_4$}
				\label{fig:d4}
			\end{subfigure}
			\hfill
			\begin{subfigure}{0.32\textwidth}
				\includegraphics[width=\textwidth]{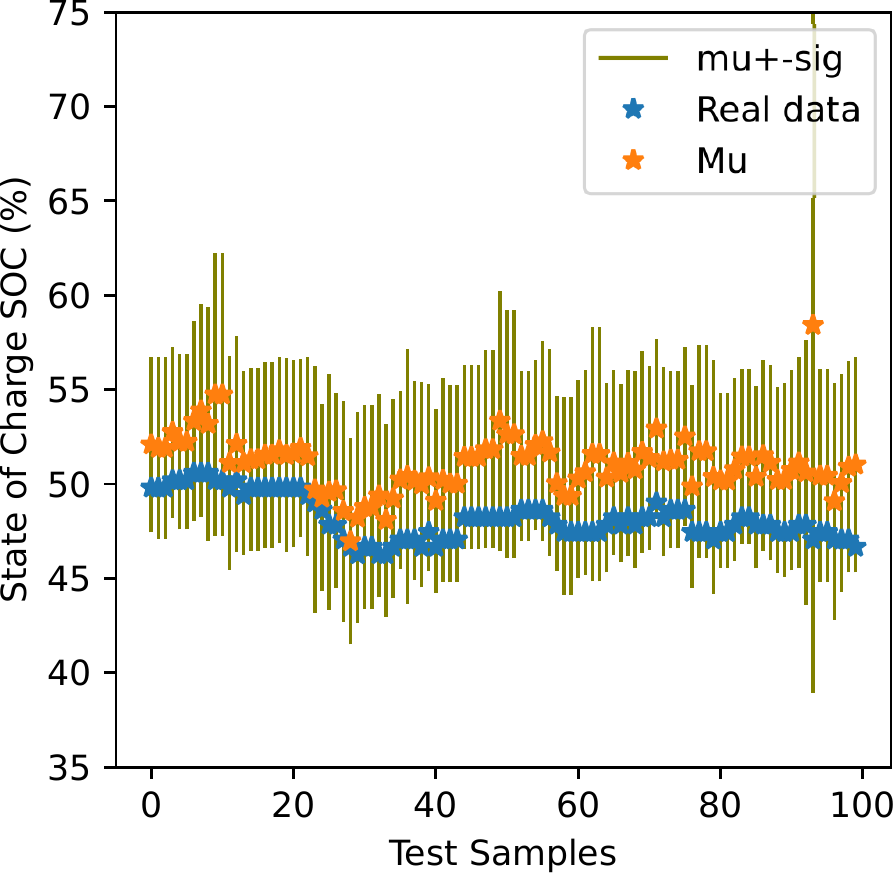}
				\caption{$D_5$}
				\label{fig:d5}
			\end{subfigure}

			\caption{Ground Truth vs. Predictions for Test Samples for Datasets $D_i, i \in [0,5]$} 
			\label{fig:results}
		\end{figure*}

		\begin{figure*}
			\label{fig:toy}
			\centering
			\begin{subfigure}{0.45\textwidth}
				
				\centering
				\includegraphics[width=\textwidth]{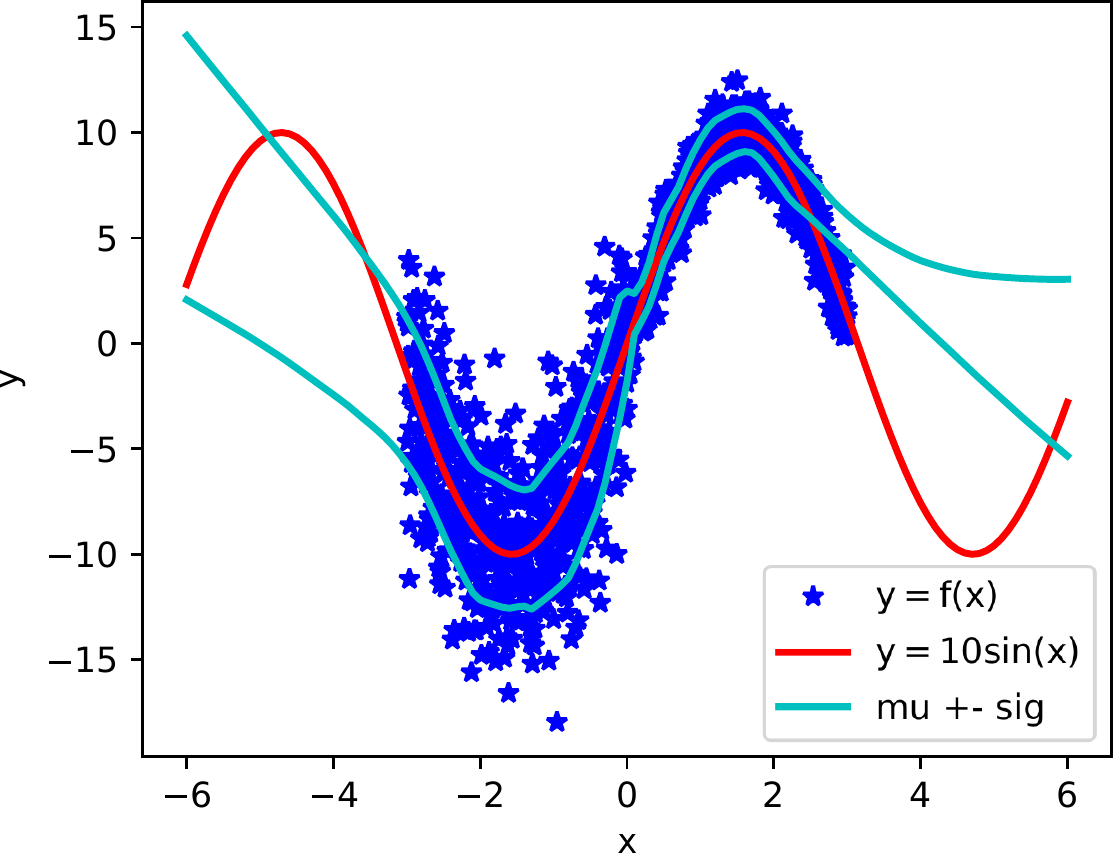}
				\caption{Comparison between ensemble and ground truth}
				\label{fig:toy1}			
			\end{subfigure}
			\begin{subfigure}{0.45\textwidth}
				
				\centering
				\includegraphics[width=\textwidth]{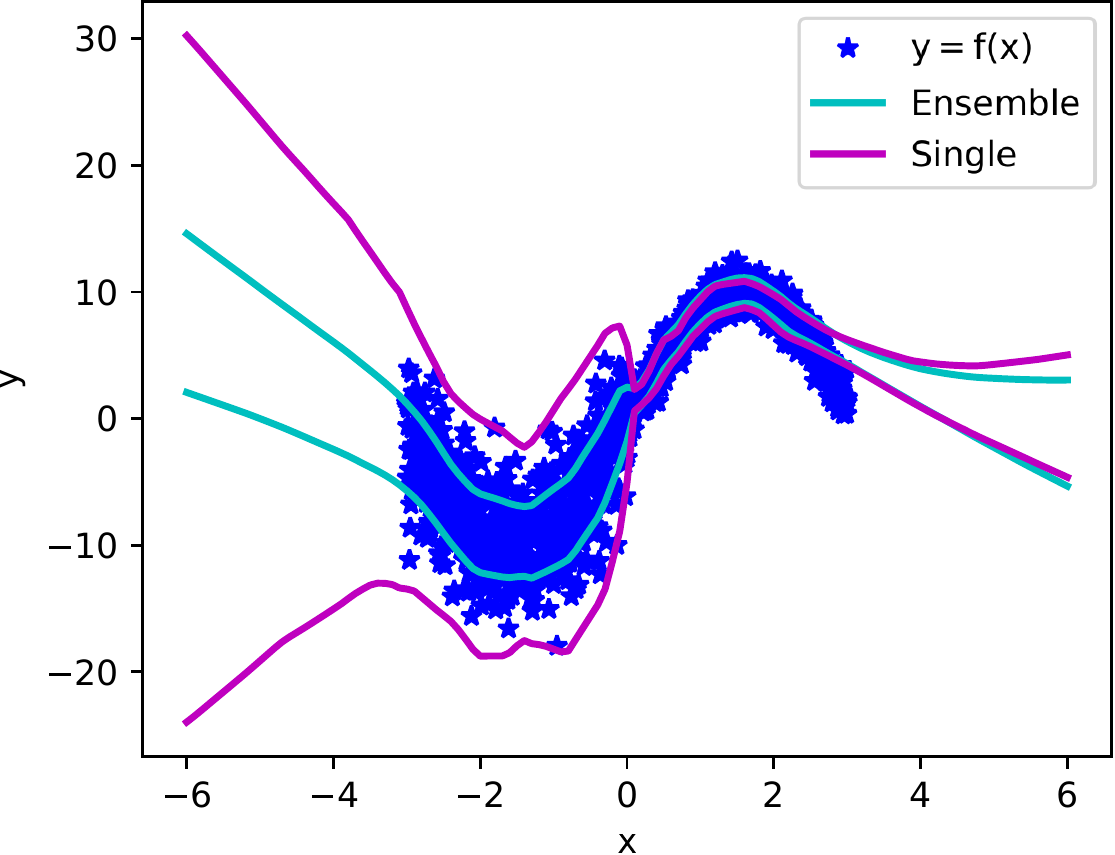}
				\caption{Comparison between ensemble and single neural network}
				\label{fig:toy2}
			\end{subfigure}
			\caption{Toy dataset results: for $x \in [-6,6]$, the real data is in blue (see Equations~\ref{eq:toy} and~\ref{eq:ep} for the $f$ expression), the ground truth in red, the ensemble's and a single network's prediction results are respectively in light blue and purple.}
			
		\end{figure*}

		\begin{table}
			\centering
			\caption{NLL and RMSE Results. The values are expressed as mean $\pm$ standard  deviation.}
			\label{table:results}
			\begin{tabular}{|c |c| c|}
				\Xhline{2\arrayrulewidth}
				\textbf{Datasets} & \textbf{NLL} & \textbf{RMSE}\\		
				\Xhline{2\arrayrulewidth}
				$D_0$ & $9.35 \pm 0.87 $ & $12.27 \pm 1.14 $\\
				$D_1$ & $7.99 \pm 0.64$ & $10.17 \pm 0.92$\\
				$D_2$ & $8.21 \pm 0.76$ & $10.69 \pm 1.02$\\
				$D_3$ & $6.72 \pm 0.49$ & $8.39 \pm 1.15$\\
				$D_4$ & $7.21 \pm 0.54$ & $8.81 \pm 1.01$\\
				$D_5$ & $6.89 \pm 0.56$ & $7.94 \pm 0.96$ \\				
				\Xhline{2\arrayrulewidth}
				
			\end{tabular}
		\end{table}
		To evaluate the method, we showed in Fig.~\ref{fig:results} the prediction results for each dataset $D_i$ alongside the ground truth data. The considered test samples are a random batch of 100 points from the testing set. For each test sample, we plot the ground-truth value of SOC and the predicted mean $\mu$, respectively in blue and orange. Furthermore, the vertical lines depict the interval $[\mu-\sigma, \mu+\sigma]$, where $\sigma$ is the predicted variance. The closer the blue and orange lines are, the more accurate and confident the prediction is. The vertical lines show the range of the predicted SOC values. Except for $D_2$, all result figures indicate that the predicted mean SOC value is within at most 5\% SOC points. As for $D_2$, while the disparity between predictions and ground truth is relatively high, it is accompanied by a higher variance range, thus indicating the reduced certainty in the prediction. Such a variance range can potentially alert the driver to the confidence level in the predictions, thus leading them to be more careful with the SOC estimations. In the other datasets, the variance interval is considerably smaller. For all datasets, the average variance interval is smaller than 15\%. \\
		
		Additionally, to demonstrate the effect of using NLL as a scoring rule, we trained ensembles with RMSE as a loss and reported the results in~\ref{table:results}. All NLL loss values are notably smaller than the RMSE counterparts, thus indicating that using NLL can help achieve high prediction accuracy as well as reduce predictive uncertainty. \\
		
		To further illustrate the results of the ensemble, we used a toy dataset comprising of only one variable $x$ and a sinusoidal output $y = f(x)$ to which a Gaussian noise is added. Equations~\ref{eq:toy} and~\ref{eq:ep} show how the real data is generated for values $x \in [-3,3]$. The ground truth data is $y=10\sin(x)$. We train the DNN ensemble following the same approach as for the $D_i$ datasets. Fig.~\ref{fig:toy1} shows the output of the ensemble, the ground truth, as well as the real data for $x \in [-6,6]$. We note that the predicted mean and variance are mostly close to the ground truth for known samples ($x \in [-3,3]$). Yet for unknown values ($x > 3 \text{ or } x < -3 $), the range of the interval $[\mu-\sigma, \mu+\sigma]$ increases to account for more uncertainty. Fig.~\ref{fig:toy2} compares the predictions of a single neural network to the ensembles'. While a single network's performance is fair for known samples, it is less smooth and consistent than that of the ensemble. As for unknown samples, we note the single network shows a bigger range than the ensemble, meaning it is associated with a higher uncertainty.

		\begin{equation}
			\label{eq:toy}
			f(x) = 10\sin(x) + \epsilon
		\end{equation}

		\begin{equation}
			\label{eq:ep}
			\epsilon =
			\begin{cases}
				\mathcal{N}(0, 9), & \text{if }  x \leq 0 \\
				\mathcal{N}(0, 1), &  \text{if }   x > 0 \\
			\end{cases}		
		\end{equation}

		\section{Discussion}
		
		\label{sec:discussion}
		The used data reflected different aspects in terms of driver behavior, weather, trajectory, terrain, and traffic conditions. Therefore, the performance's stability throughout the six datasets indicates a robustness of the approach and its capability to adapt. The utilization of an ensemble with the the negative log likelihook as a scoring rule was crucial to obtain these results. Furthermore, the randomization of batches when training was beneficial because it exposed the learning algorithm to more diversity and variance, hence mitigating overfitting.
		
		While it is common to construct ensembles of numerous predictors (possibly hundreds), we show that limiting the ensemble size to five neural networks is sufficient and it has the benefit of lowering the computation time. On the one hand, the comparison with RMSE shows a considerable benefit to using NLL as shown in Table~\ref{table:results}. Not only are the mean values for NLL smaller than those for RMSE, we also note that the same goes for the standard deviations. On the other hand, the comparison between a single neural network and ensemble for the toy dataset (in Fig.~\ref{fig:toy1}) demonstrates the considerable improvement an ensemble can entail. The range of predicted values for the ensemble remain smaller than those for the single neural network even for unknown values. However, the fact that this range increases in both cases for unknown input values shows that the model is sensitive to new input and provides less confident predictions than for input it previously encountered. This aspect can be informative to the driver, meaning that if the predicted variance is high, then they should be cautious about the interpretation of the predicted value for the battery's state of charge. In a critical and dynamic activity such as driving, it is useful to have such a model that reliably predicts the battery's state of charge and where the user is informed about the estimation's confidence.

		\section{Conclusion and Future Work}
		
		\label{sec:conclusion}
		We describe in this paper a data-driven method aiming to predict the state of charge of a hybrid vehicle battery. It consisted of a deep neural network ensemble trained to minimize the negative log likelihood; a loss that targets predictive uncertainty. Given the multiple sources of variability that can affect the energy consumption of battery-equipped vehicles, this method can be beneficial, since it accounts for uncertainty and returns mean and variance of predictions. This output can be utilized online to continuously inform the driver about the battery state, or offline to analyze the energy consumption of a vehicle and extract relevant insights. Since it relies on commonly available battery attributes, this method can be easily reproduced and would help contribute to higher usability of battery equipped vehicles, by boosting the driver's trust in the battery state output.
				 
		As a future step, we will consider making the method more lightweight to further facilitate its online deployment on different platforms that might have less computation resources available. For instance, we will explore the approach of sharing parameters between the ensemble components to accelerate training. We will also investigate the use of data from different vehicles in more various driving conditions. \\
		
		\section*{Acknowledgment}

		This work was supported by the Austrian Ministry for Climate Action, Environment, Energy, Mobility, Innovation and Technology (BMK) Endowed Professorship for Sustainable Transport Logistics 4.0., IAV France S.A.S.U., IAV GmbH, Austrian Post AG and the UAS Technikum Wien. \\
		
		\bibliographystyle{IEEEtran}
		\bibliography{jk_jabref}
		
	\end{document}